\documentclass{article}
\usepackage{ijcai21}

\usepackage{times}
\usepackage{soul}
\usepackage{url}
\usepackage[hidelinks]{hyperref}
\usepackage[utf8]{inputenc}
\usepackage[small]{caption}
\usepackage{graphicx}
\usepackage{amsmath}
\usepackage{amsthm}
\usepackage{booktabs}
\usepackage{algorithm}
\usepackage{epsfig}
\usepackage{amssymb}
\usepackage{dsfont}
\usepackage{float}
\title{VR3Dense: Voxel Representation Learning for 3D Object Detection and Monocular Dense Depth Reconstruction}

\author{
    Shubham Shrivastava\thanks{Work done as a graduate student at Stanford University}
    \affiliations
    {Ford Greenfield Labs, Palo Alto, CA}
    \emails
    {sshriva5@ford.com}
}

\begin{document}

\maketitle

\begin{abstract}
3D object detection and dense depth estimation are one of the most vital tasks in autonomous driving. Multiple sensor modalities can jointly attribute towards better robot perception, and to that end, we introduce a method for jointly training 3D object detection and monocular dense depth reconstruction neural networks. It takes as inputs, a LiDAR point-cloud, and a single RGB image during inference and produces object pose predictions as well as a densely reconstructed depth map. LiDAR point-cloud is converted into a set of voxels, and its features are extracted using 3D convolution layers, from which we regress object pose parameters. Corresponding RGB image features are extracted using another 2D convolutional neural network. We further use these combined features to predict a dense depth map. While our object detection is trained in a supervised manner, the depth prediction network is trained with both self-supervised and supervised loss functions. We also introduce a loss function, edge-preserving smooth loss, and show that this results in better depth estimation compared to the edge-aware smooth loss function, frequently used in depth prediction works. 
\footnote{Code implementation available here: \url{https://github.com/towardsautonomy/VR3Dense}}
\end{abstract}

%%%%%%%%% BODY TEXT
\section{Introduction}\label{sec:intro}

\begin{figure}[!ht]
\centering
  \includegraphics[width=0.85\linewidth]{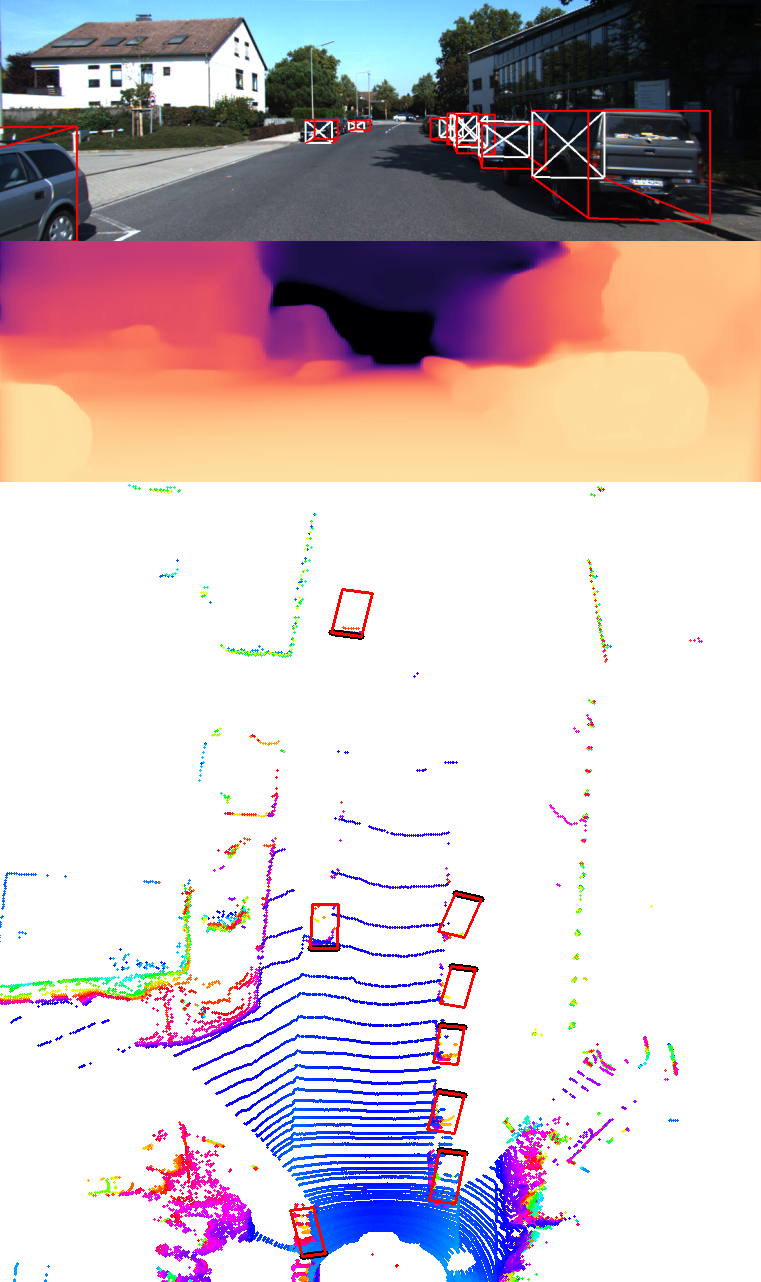}
  \caption{Joint prediction of object poses and dense depth map from a LiDAR point-cloud and an RGB image.}
  \label{fig:intro_fig}
  \vspace{-0.2cm}
\end{figure}

3D Object Detection is one of the most significant part in autonomous vehicle perception. An autonomous vehicle needs to be aware of its surrounding objects and should be capable of predicting their future trajectory. Most autonomous vehicles in their development phase today are occupied with LiDARs, Cameras, and RADARs which allow them to perceive the environment. While 2D object detection methods using cameras \cite{redmon2016look,tan2020efficientdet} have matured quite a bit, it does not really capture the accurate geometry of the scene and hence does not help an autonomous vehicle plan its actions. Few recent papers have also proposed the use of monocular camera to detect object 6-DoF poses \cite{shrivastava2021cubifae3d,brazil2019m3d,qin2019monogrnet}, however, these have proven to be inferior to LiDAR based object detection methods. LiDARs have been one of the most crucial sensors for obtaining accurate 3D scene representation, however, its high-cost and sparsity in the point-cloud data have discouraged a small set of automotive companies from utilizing it for perception tasks. Although expensive, it is irrefutable that LiDAR point-cloud provides means of obtaining the most accurate 3D geometry of the scene. 

Monocular depth estimation is another field of computer vision research that has been getting a lot of attention recently in 3D scene understanding. Obtaining pixel-wise depth map provides us with a plethora of information about the scene and helps machine understand a dense representation of the environment. A major deterrent in obtaining dense depth though is the scarcity of available public datasets and the hardships involved in generating ground-truth labels. Most available dense depth estimation datasets like NYU Depth Dataset V2 \cite{Silberman:ECCV12} uses RGB-D cameras for indoor settings, and classical computer vision techniques such as stereo matching, LiDAR points super-resolution \cite{Uhrig2017THREEDV}, or LiDAR points projected onto the image plane \cite{packnet} for the outdoor environment. This has encouraged the research community to innovate several self-supervised and semi-supervised methods for RGB image to depth map estimation \cite{monodepth17,packnet,semisupmonodepth19}.

We combine the two streams of perception tasks and introduce a method of jointly training LiDAR point-cloud based 3D object detection and monocular image to dense depth estimation neural networks. We represent the input point-cloud as a set of non-cubic voxels, each encoding the density of points contained within, and then use a set of 3D convolution layers for extracting spatial features which are concatenated with the latent vector of RGB image to depth estimation network. We pose 3D object detection as a regression problem and regress object pose parameters along with their confidence and class probabilities from the extracted spatial features using a set of fully-connected layers. The dense depth estimation network on the other hand is an hourglass architecture with skip connections, built using 2D convolution layers. We call our method - VR3Dense, which requires point-cloud, left and right stereo images, and object pose labels for training. During inference, we only require a LiDAR point-cloud and corresponding left camera image as inputs and predict object poses along with pixel-wise dense depth map as outputs. We work with KITTI 3D object detection dataset \cite{Geiger2012CVPR} for training and testing our method.

\section{Related Work}

Several LiDAR-based 3D object detection methods have proven to perform extremely well and achieve state-of-the-art results on existing dataset benchmarks. Researchers utilize various representations of point-clouds such as bird-eye view (BEV), voxels, and stixels - each with its own advantages and disadvantages. While methods such as \cite{li2016vehicle}, and PIXOR \cite{yang2019pixor} projects LiDAR point-cloud on a 2D plane and then use 2D CNN for object detection, other methods such as Vote3deep \cite{engelcke2017vote3deep} and Voxelnet \cite{zhou2017voxelnet} represents point-cloud as a set of 3D voxels and then use 3D convolution to extract features. These methods often employ separate heads for objectness probability prediction and pose prediction. Another stream of research such as VoxNet \cite{Maturana2015VoxNet}, converts a point-cloud segment into a 3D volumetric grid and then perform detection. Recently, PointNet \cite{qi2017pointnet} introduced a different approach to point-representation which eliminates the need to manually structure the points in a predefined manner and was further improved by PointNet++ \cite{qi2017pointnetpp}.

Due to scantiness of paired image and depth training datasets, several self-supervised and semi-supervised techniques have been proposed. All these methods use more than one views and poses between them during training to reproject pixels across frames and compute reprojection losses which provides supervision to the network. SfMLearner \cite{zhou2017unsupervised} requires only monocular video sequences during training. It predict poses between multi-view frames using a \textit{Pose CNN} and predicts a depth map for the target view using \textit{Depth CNN}. Other views are then warped to the target view and the reprojection loss thus computed helps both the network to be trained in an end-to-end fashion. PackNet-SfM \cite{packnet} follows a similar path but leverages the camera’s velocity when available to solve for the scale ambiguity with monocular vision. They also use \textit{3D packing} and \textit{unpacking} blocks in their encoder and decoder to preserve high resolution depth decoding. Similarly, Monodepth \cite{monodepth17} uses left and right stereo images with known baseline and exploits epipolar geometry to warp each image to the other in order to compute left-right reprojection losses along with disparity consistency loss which feeds supervisory signal to the network during training. Few other works such as \cite{semisupervisedmonodepth177} take a semi-supervised approach, where they also use LiDAR points projected on to the image plane to add supervision to the pixels for which a 3D point projection is available.

We take inspiration from all these work to build a 3D object detection network and a semi-supervised monocular depth estimation network. Our object detection network uses 3D convolution layers whereas our depth estimation network uses 2D convolutions; their representations are merged downstream from which our depth estimation decoder benefits and provides a better quality depth reconstruction.\vspace{-0.05cm}

\section{Technical Approach}\label{sec:technical_approach}

\begin{figure*}[!ht]
\centering
  \includegraphics[width=0.95\linewidth]{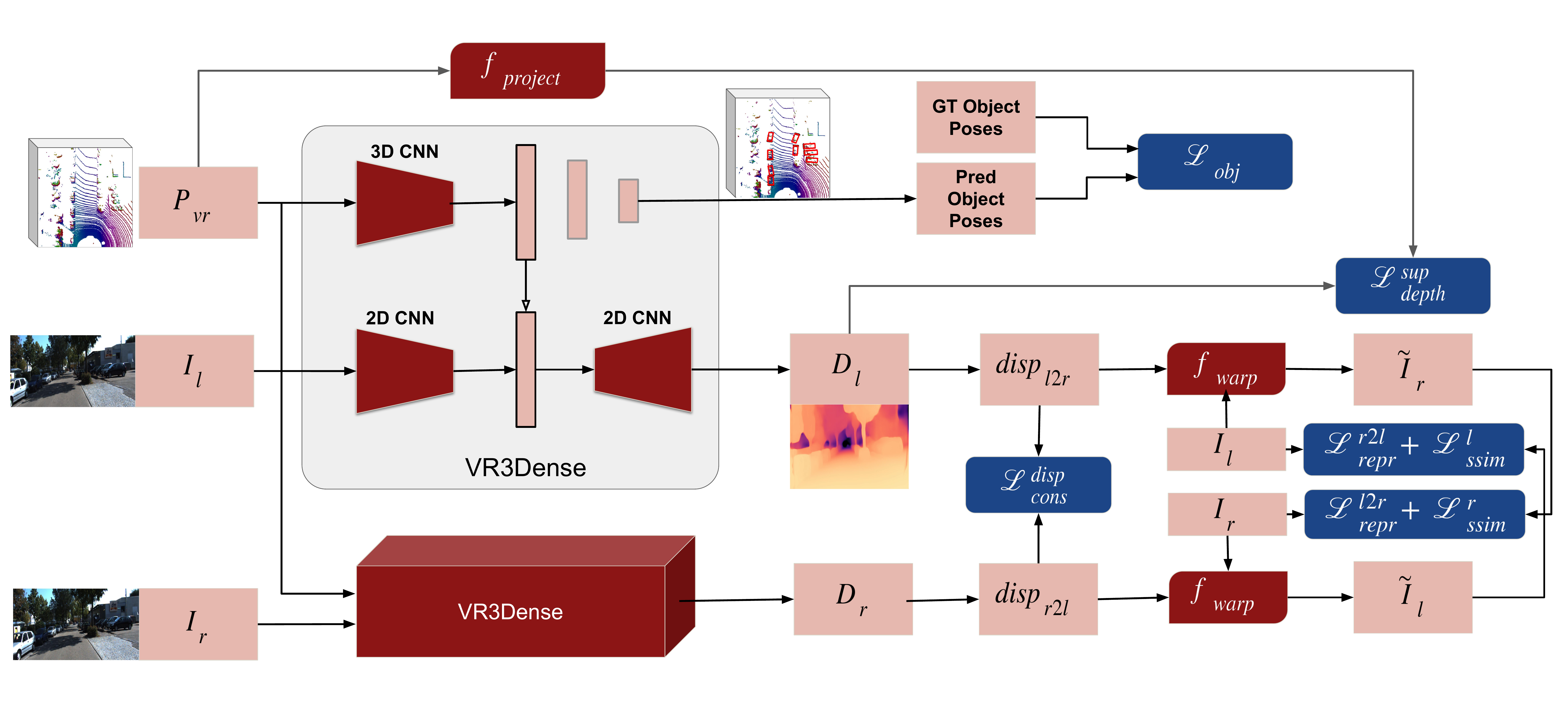}
  \caption{VR3Dense jointly learns 3D object detection from LiDAR point-cloud and dense depth estimation from corresponding monocular camera frame. 3D object detection is trained with supervision from ground-truth labels while monocular depth estimation is trained in a semi-supervised way; self-supervision from stereo image reconstruction and supervision from sparse LiDAR points projected onto the image plane. Learned feature vector from voxel representation is concatenated with the latent vector of depth estimation network which allows depth decoder to take advantage of spatial features extracted from LiDAR points.}
  \label{fig:vr3dense_approach}
  \vspace{-0.2cm}
\end{figure*}

Learning representations from more than one sensor modality furnishes a richer set of information and helps solve perception tasks more accurately. Figure \ref{fig:vr3dense_approach} summarizes our technical approach for joint 3D object detection and monocular depth estimation. For learning spatial representation from LiDAR point-clouds, we first encode points into a collection of non-cubic voxels, $P_{vr}$, with each voxel encoding density of points within the occupied volume. This voxelized point-cloud is then passed to a 3D convolutional neural network for feature extraction. At the same time, corresponding left stereo camera image, $I_l$ is passed through a U-Net \cite{unet2015} like encoder-decoder architecture with residual blocks to produce a dense depth map, $D_l$. Encoder of this network encodes the RGB image into a latent vector, to which the encoded voxelized point-cloud features are concatenated before being fed to the decoder. We qualitatively and quantitatively show that this results in better depth estimation. We further use fully-connected layers to extract object pose parameter along with class probabilities from encoded voxel representations.

\subsection{LiDAR point-cloud based 3D Object Detection}

For 3D object detection, we divide the voxelized region-of-interest into a set of volumes, each tasked with predicting an \textit{objectness score} and the corresponding \textit{object pose}. Since, a prior knowledge of the environment is known (e.g. cars mostly drive on the road - same plane as ego vehicle), we assume the roll and pitch to be zero. Additionally, class probabilities are also predicted as a one-hot vector for object classification - $(p_{class1}, p_{class2}, ..., p_{classN})$. This results in the predicted output pose vector to be of size $M \times (9 + N)$ for a maximum of $M$ objects with prediction for each object corresponding to ($confidence$, $x_{center}$, $y_{center}$, $z_{center}$, $length$, $width$, $height$, $cos({\theta})$, $sin({\theta})$, $p_{class1}$, $p_{class2}$, ..., $p_{classN}$), where $\theta$ is $yaw$ angle of the object relative to ego-vehicle. We minimize $L2$ loss for computing $x_{center}$, $y_{center}$, $z_{center}$, $length$, $width$, $height$, $cos(\theta)$, and $sin(\theta)$ errors as shown in equation \ref{eqn:pose_loss} and optimize for their minimization using $Adam$ optimizer. 

\begin{equation}\label{eqn:pose_loss}
    \mathcal{L}_{pose} = \mathds{1}_{obj} \frac{\sum{|| y_{pose} - \hat{y}_{pose} ||_2}}{n_{true\_obj} + \epsilon}
\end{equation}

Where, $pose$ $\in$ [$x_{center}$, $y_{center}$, $z_{center}$, $length$, $width$, $height$, $cos(\theta)$, $sin(\theta)$].

For confidence loss computation, we again use $L2$-norm, but weigh them based on number of positive and negative objects in ground-truth labels as given in equation \ref{eqn:conf_loss}.

\begin{equation}\label{eqn:conf_loss}
\begin{split}
    \mathcal{L}_{conf} & = \mathds{1}_{obj} \frac{\sum{|| y_{conf} - \hat{y}_{conf} ||_2}}{n_{true\_obj} + \epsilon}  + \\
    & (1 - \mathds{1}_{obj}) \frac{\sum{|| y_{conf} - \hat{y}_{conf} ||_2}}{n_{false\_obj} + \epsilon}
\end{split}
\end{equation}

We use cross-entropy loss for the optimization of object class prediction ($\mathcal{L}_{class}$). Additionally, we attempt to maximize generalized-intersection-over-union (GIoU) \cite{rezatofighi2019generalized} directly during the optimization process by minimizing the GIoU loss given in equation \ref{eqn:giou_loss}.

\begin{equation}\label{eqn:giou_loss}
    \mathcal{L}_{giou} = || GIoU(y, \hat{y}) - \mathds{1}_{obj}||_2
\end{equation}

In equations \ref{eqn:pose_loss}, \ref{eqn:conf_loss}, and \ref{eqn:giou_loss}, $\mathds{1}_{obj}$ is $1$ if an object exist in the ground-truth label vector, and $0$ otherwise. Our complete object detection loss function is a weighted sum of these individual loss terms and is given in equation \ref{eqn:obj_det_loss}, where $\lambda_n$ are loss weights.

\begin{equation}\label{eqn:obj_det_loss}
    \mathcal{L}_{obj} = \lambda_1 \mathcal{L}_{conf} + \lambda_2 \mathcal{L}_{pose} + \lambda_3 \mathcal{L}_{class} + \lambda_4 \mathcal{L}_{giou}
\end{equation}

\subsection{Semi-Supervised Monocular Depth Estimation}

\cite{semisupervisedmonodepth177} and \cite{semisupmonodepth19} showed that semi-supervised monocular depth estimation outperforms both supervised and unsupervised methods. Following these papers, we train our depth estimation network with supervision from LiDAR points projected onto the image plane in addition to self-supervision from stereo image reconstruction losses. Unlike other papers however, we attempt to directly predict a higher resolution depth map rather than coarse-to-fine disparity or inverse depth. We notice that the loss functions used in most methods alone are not sufficient for producing high-resolution images with sharp edges around object silhouettes. State-of-the-art monocular depth estimation methods use \textit{edge-aware smooth loss} as given in equation \ref{eqn:edge_aware_smooth_loss}, which penalizes gradients in depth map if a corresponding gradient does not exist in the RGB image. It however, does not penalize the network if predicted depth map is too smooth, even across object boundaries - which is precisely the reason for the predicted depth map not being sharp at full resolution when a coarse-to-fine approach is not taken. To this end, we propose a loss function called \textit{edge perseverance loss}, which promotes edges in depth map proportional to the edges in RGB image and is given in equation \ref{eqn:edge_preserv_loss}. Since the magnitude of depth gradients and image gradients might not be of the same scale, we attempt to learn this proportionality parameter ($\alpha_0$ and $\alpha_1$ in equation \ref{eqn:edge_preserv_loss}). We use a combination thereof which we call - \textit{edge-preserving smooth loss}, during our monocular depth estimation training and is given in equation \ref{eqn:edge_preserv_smooth_loss}.

\begin{equation}\label{eqn:edge_aware_smooth_loss}
    \mathcal{L}_{smooth} = \frac{1}{N}\sum_{n \in N}{|\partial_x \hat{D}|e^{-|\partial_x I|} + |\partial_y \hat{D}|e^{-|\partial_y I|}}
\end{equation}

Where, $\hat{D}$ is the predicted depth map, $I$ is the input image, and $N$ is total number of pixels. 

\begin{equation}\label{eqn:edge_preserv_loss}
    \mathcal{L}_{ep} = \frac{1}{N}\sum_{n \in N}{\frac{e^{|\partial_x \hat{D} - \alpha_0 \partial_x I|}+ e^{|\partial_y \hat{D} - \alpha_1 \partial_y I|}}{2}} - 1
\end{equation}

\begin{equation}\label{eqn:edge_preserv_smooth_loss}
    \mathcal{L}_{eps} = \beta_{edge} \mathcal{L}_{ep} + (1 - \beta_{edge}) \mathcal{L}_{smooth}
\end{equation}

In equation \ref{eqn:edge_preserv_loss}, $\alpha_0 = tanh(W_0 \partial_x I + b_0)$ and $\alpha_1 = tanh(W_1 \partial_x I + b_1)$, where $W_n$ and $b_n$ are learnable parameters for $n \in [0,1]$. $\beta_{edge}$ shown in equation \ref{eqn:edge_preserv_smooth_loss} is a hyperparameter and we set it to $0.5$ during our experiments.

For self-supervision, we use stereo image reconstruction to compute reprojection losses by warping left stereo image ($I_l$) to right ($\tilde{I}_r$) and right image ($I_r$) to left ($\tilde{I}_l$). We do this by first converting predicted depth from the left image ($D_l$) into disparity ($disp_{l2r}$) and then use PyTorch's \cite{NEURIPS2019_9015} differentiable bilinear sampling ($f_{warp}$) to warp the left stereo image to right as shown in equation \ref{eqn:displ2r}. 

\begin{equation}\label{eqn:displ2r}
\begin{split}
    disp_{l2r} & = \frac{fb}{D_l} \\
    \tilde{I}_r & = f_{warp}(I_l, disp_{l2r})
\end{split}
\end{equation}

Similarly, we predict the depth map ($D_r$) for right stereo image and then warp it to the left using right-to-left disparity, which can be computed as $disp_{r2l} = -fb/D_r$. Stereo image reprojection losses are then computed using equation \ref{eqn:reproj_err}, where $\mathcal{M}_{l2r}$ and $\mathcal{M}_{r2l}$ represents a mask of all \textit{ones} warped from left-to-right using $disp_{l2r}$, and right-to-left using $disp_{r2l}$ respectively; and $\odot$ represents an element-wise multiplication. One could also incorporate an additional reprojection loss which measures the consistency between left image and right-to-left warped image using the negative of $disp_{l2r}$, and vice versa.

\begin{equation}\label{eqn:reproj_err}
\begin{split}
    \mathcal{L}_{repr}^{l2r} & = \frac{1}{N}\sum_{n \in N}{\mathcal{M}_{l2r} \odot Huber(I_r, \tilde{I}_r)} \\
    \mathcal{L}_{repr}^{r2l} & = \frac{1}{N}\sum_{n \in N}{\mathcal{M}_{r2l} \odot Huber(I_l, \tilde{I}_l)} \\
    \mathcal{L}_{repr} & = \frac{\mathcal{L}_{repr}^{l2r} + \mathcal{L}_{repr}^{r2l}}{2} 
\end{split}
\end{equation}

We also enforce consistency between the disparity computed from left and right images and is given by $\mathcal{L}_{cons}^{disp}$ in equation \ref{eqn:disp_consist_loss}. We also compute \textit{structure similarity index} between one image and its reconstructed version from the other image as shown in equation \ref{eqn:app_match_loss} which gets incorporated into the \textit{appearance matching loss}, $\mathcal{L}_{app}$. Furthermore, we use \textit{Huber}-norm instead of commonly used \textit{L1}-norm for computing reprojection and consistency errors - we find it to converge faster in our experiments. 
    
\begin{equation}\label{eqn:disp_consist_loss}
    \mathcal{L}_{cons}^{disp} = \frac{1}{N}\sum_{n \in N}{Huber(disp_{l2r}, -disp_{r2l})} 
\end{equation}

\begin{equation}\label{eqn:app_match_loss}
\begin{split}
    \mathcal{L}_{ssim} & = \frac{1}{N}\sum_{n \in N}{\frac{1}{4}\Big(2 - 
                      \big[SSIM(I_l, \tilde{I}_l) 
                         + SSIM(I_r, \tilde{I}_r)\big]\Big)} \\
    \mathcal{L}_{app} & = \alpha_{ssim} \mathcal{L}_{ssim} + (1 - \alpha_{ssim}) \mathcal{L}_{repr} \\
\end{split}
\end{equation}

Loss functions given in equations \ref{eqn:edge_preserv_smooth_loss}, \ref{eqn:reproj_err}, \ref{eqn:disp_consist_loss}, and \ref{eqn:app_match_loss} together devise our unsupervised loss function $\mathcal{L}_{depth}^{unsup}$.

\begin{equation}\label{eqn:depth_unsup_loss}
    \mathcal{L}_{depth}^{unsup} = \lambda_5 \mathcal{L}_{eps} + \lambda_6 \mathcal{L}_{repr} + \lambda_7 \mathcal{L}_{cons}^{disp} + \lambda_8 \mathcal{L}_{app}
\end{equation}

Semi-supervision in our depth estimation network comes from the 2D projection of LiDAR points. We use an \textit{L2} loss between our depth prediction and the projected LiDAR points as given in equation \ref{eqn:depth_sup_loss}, where $P$ represents LiDAR points, $f_{project}()$ is a 2D projection function which projects LiDAR point-cloud onto the left camera image plane, $K$ is the number points projected within the image boundary, and $\mathcal{M}_{proj}$ is a mask which is $1$ at pixel locations for which we have a corresponding projected LiDAR point available and $0$ otherwise.

\begin{equation}\label{eqn:depth_sup_loss}
    \mathcal{L}_{depth}^{sup} = \lambda_9 \frac{1}{K}\sum_{k \in K}{\mathcal{M}_{proj} \odot ||f_{project}(P) - D_l ||_2}
\end{equation}

We also find that the supervised loss, $\mathcal{L}_{depth}^{sup}$, is helpful during initial parts of the training for faster convergence but does not help much later in the training process. Hence, we use a decay rate of $0.01$ for this loss during training.

\section{Experimental Section}

For encoding the LiDAR point-cloud into a set of voxels, we take the points within a region-of-interest ($0.0$ to $70.0$ meters in $x$-dimension, $-25.0$ to $25.0$ meters in $y$-dimension, and $-2.5$ to $1.0$ meters in $z$-direction) and convert them into $256 \times 256 \times 16$ non-cubic voxels, where each voxel also encode the number of points within its volume. In our coordinate convention, $x$ points ahead, $y$ points to the left of the ego-vehicle, and $z$ points up. For the purpose of predicting object presence, their poses, and the classification one-hot vector, we further divide our voxelized space into $xy$ grids of size $16$x$16$, and allow each grid to predict up to $1$ object. Images are resized to a resolution of $512 \times 256$ before passing them through the network. Our predicted depth resolution is also of the same resolution, and we up-sample it to full resolution using \textit{nearest neighbor interpolation} before computing losses for depth estimation.

\subsection{KITTI Dataset}

We use KITTI 3D object $training$ dataset \cite{Geiger2012CVPR} to train both 3D object detection and dense depth reconstruction. This dataset contains $7481$ \textit{training} samples and $7518$ \textit{testing} samples (test split). Following \cite{Shi_2019_CVPR}, we divide the training samples into \textit{train} split (3712 samples) and \textit{val} split (3769  samples) for evaluation of our object detection results. For our depth estimation training, we use \textit{training} dataset of $7518$ samples, and the entire object \textit{testing} dataset with $7518$ samples for evaluation. 

\subsection{3D Object Detection}

As summarized in table \ref{tab:quant_results_object}, our evaluation uses \textit{mean average precision} with $40$-recall points. For an IOU of $0.7$, our method does not perform as well as state-of-the-art methods. For lower IOU however, ($0.3$, and $0.7$), our method achieves high $mAP$. This illustrates that VR3Dense predictions are accurate in terms of predicting object location, however it lacks the ability of predicting a very \textit{tight} object boundary and orientation. We hypothesize that, since we use a sparse encoding of point-cloud, the network is not able to find accurate object corners. This issue can be alleviated to some extent by encoding some scene prior into the network such as anchor boxes - and will be the subject of our future work. 

\begin{table}[!htb]
    \begin{center}
      \caption{Performance on the KITTI \textit{val} split set with mAP calculated by 40 recall positions for car class. $mAP_x$ identify the \textit{Mean Average Precision (mAP)} for a 3D IOU threshold of $x$. Higher is better.}
      \label{tab:quant_results_object}
      \begin{tabular}{l||l|l|l}
      \toprule
        \textbf{Method}&$mAP_{\textbf{.3}}$&$mAP_{{\textbf{.5}}}$&$mAP_{\textbf{.7}}$ \\
        \midrule
        \cite{zhou2017voxelnet} & - & - & 70.1 \\
        \cite{Shi_2019_CVPR} & - & - & 74.8 \\
        \cite{shi2020pv} & - & - & 85.5 \\
        Ours    &   96.1    &   85.3   &   36.2 \\
        \bottomrule
      \end{tabular}
  \end{center}
\end{table}

\subsection{Semi-Supervised Monocular Depth Estimation}

\begin{table*}[!htb]
    \begin{center}
      \caption{Monocular Depth Estimation results on KITTI object \textit{test} dataset. Evaluations are done on the pixels for which we have projected points from LiDAR available.}
      \label{tab:quant_results_depth}
      \begin{tabular}{l|l||l|l|l|l|l|l|l}
      \toprule
        \textbf{Method} & Range & Abs Rel & Sq Rel & RMSE & RMSE log & $\delta < 1.25$ & $\delta < 1.25^2$ & $\delta < 1.25^3$ \\
        \midrule
         & & \multicolumn{4}{c|}{\textit{Lower is better}} & \multicolumn{3}{c}{\textit{Higher is better}}\\ 
        \midrule
        \midrule
        \cite{zhou2017unsupervised} & 0-50m & 0.201 & 1.391 & 5.181 & 0.264 & 0.696 & 0.900 & 0.966 \\
        \cite{gargetal} & 0-50m & 0.169 & 1.080 & 5.104 & 0.273 & 0.740 & 0.904 & 0.962 \\
        \cite{monodepth17} & 0-50m & 0.140 & 0.976 & 4.471 & 0.232 & 0.818 & 0.931 & 0.969 \\
        \cite{semisupervisedmonodepth177} & 1-50m & - & - & 3.518 & 0.179 & 0.875 & 0.964 & 0.988 \\
        \cite{semisupmonodepth19} & 0-80m & 0.096 & 0.552 & 3.995 & 0.152 & 0.892 & 0.972 & 0.992 \\
        Ours  & 0-30m & 0.214432 & 3.724554 & 5.561154 & 0.287028 & 0.788234 & 0.901136 & 0.950242 \\
        Ours  & 0-50m & 0.217422 & 3.775980 & 6.507308 & 0.296695 & 0.769214 & 0.893053 & 0.946789 \\
        Ours  & 0-70m & 0.219240 & 3.859928 & 7.164907 & 0.302273 & 0.760768 & 0.888690 & 0.944800 \\
        \bottomrule
      \end{tabular}
  \end{center}
\end{table*}

For our depth estimation evaluation, we use $Abs\ Rel$, $Sq\ Rel$, $RMSE$, $RMSE\ log$, $\delta < 1.25$, $\delta < 1.25^2$, and $\delta < 1.25^3$ to compute performance metrics considering the LiDAR points projected onto image plane as ground-truth. For depth estimation, we split the KITTI 3D object detection \textit{training} dataset into $90\%$ \textit{train} and $10\%$ \textit{val} split. Quantitative results of our semi-supervised monocular depth estimation is summarized in table \ref{tab:quant_results_depth}.

\subsection{Joint training of 3D object detection and monocular depth estimation}

\begin{figure*}[!ht]
  \centering 
  \includegraphics[width=0.24\linewidth]{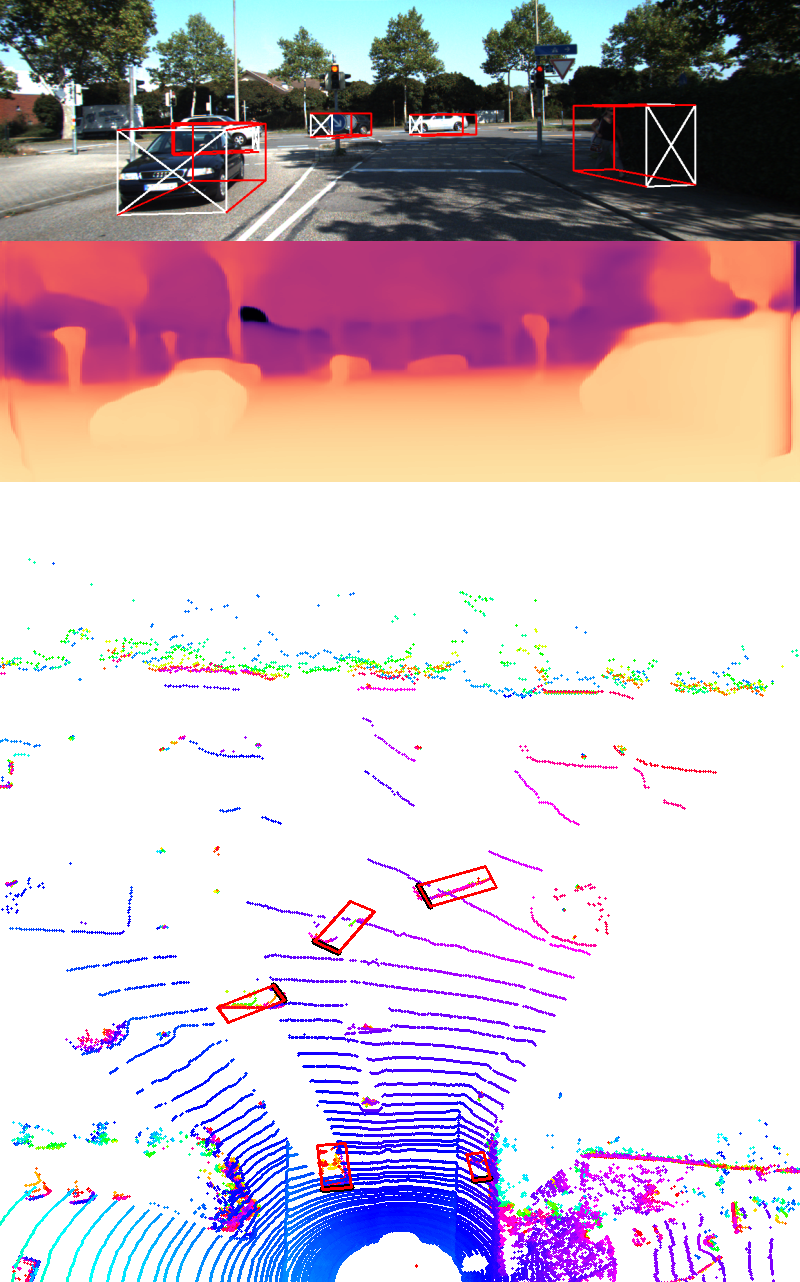}
  \includegraphics[width=0.24\linewidth]{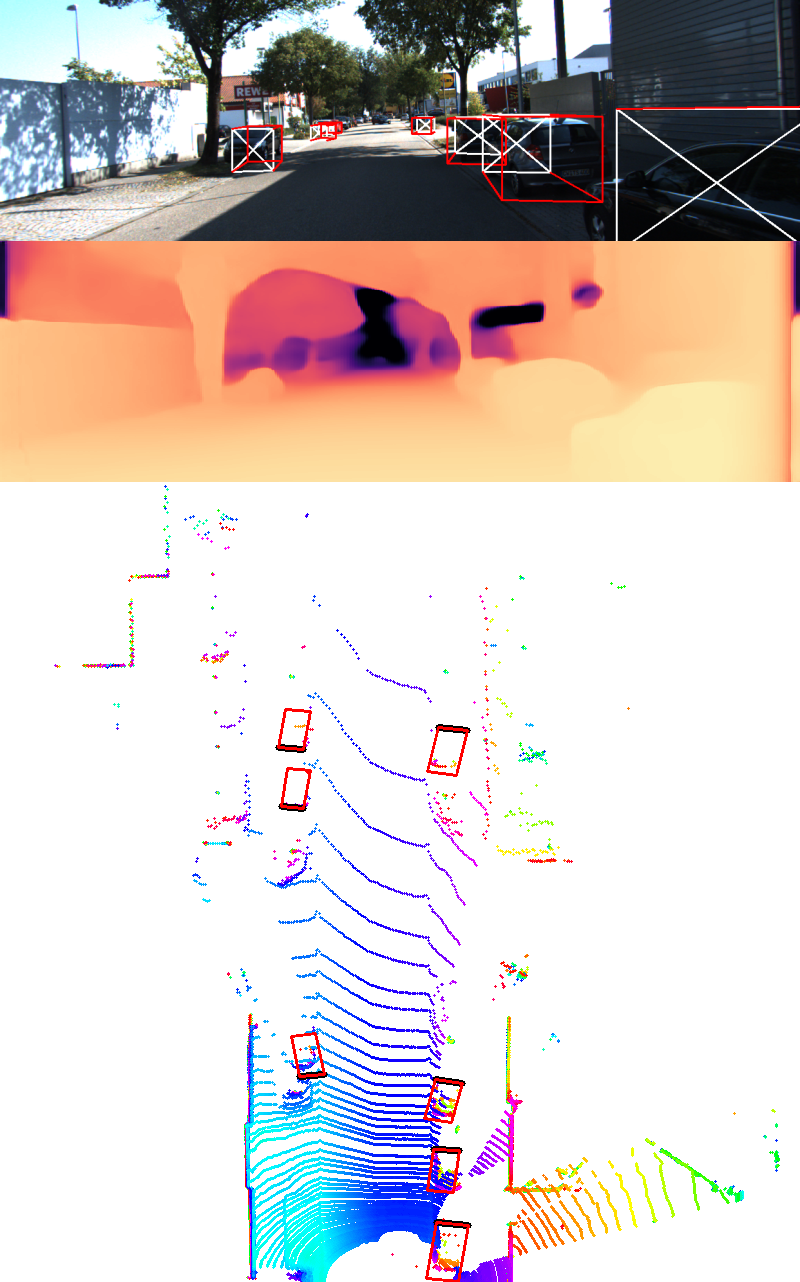}
  \includegraphics[width=0.24\linewidth]{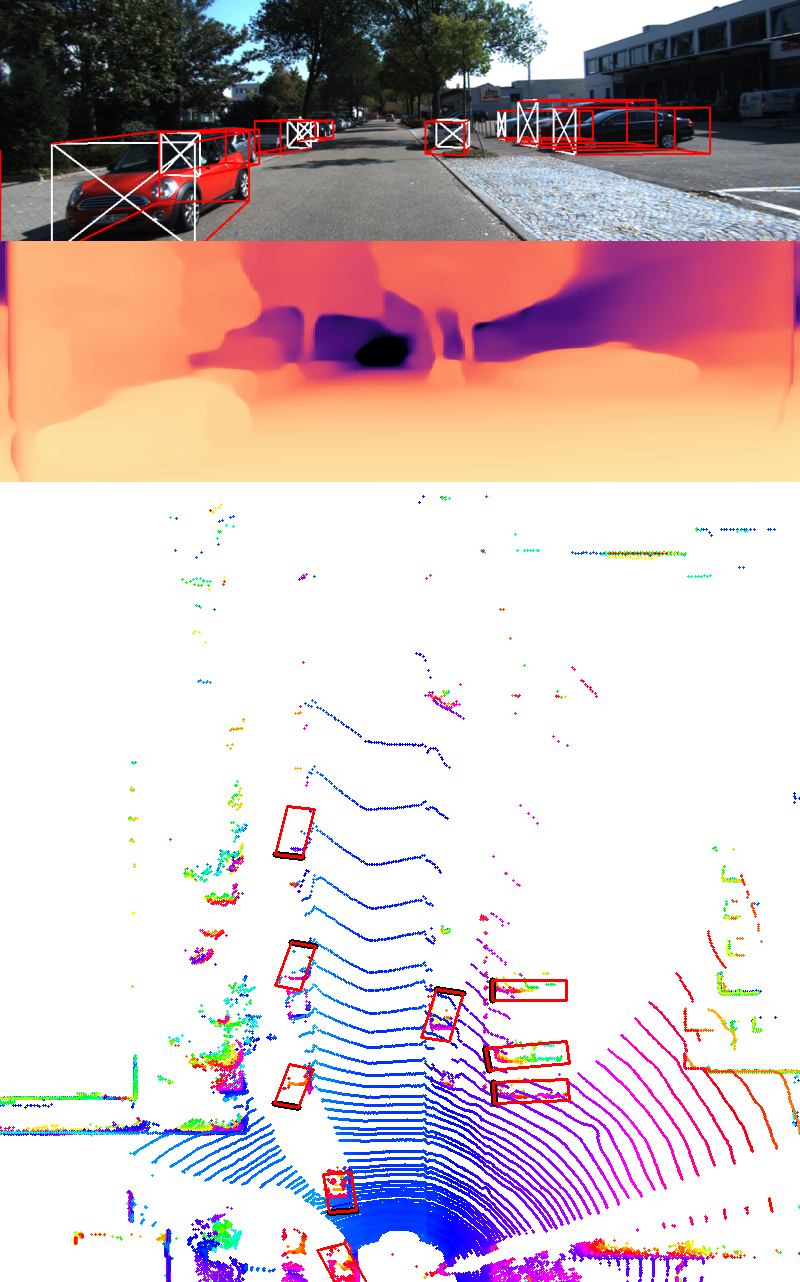}
  \includegraphics[width=0.24\linewidth]{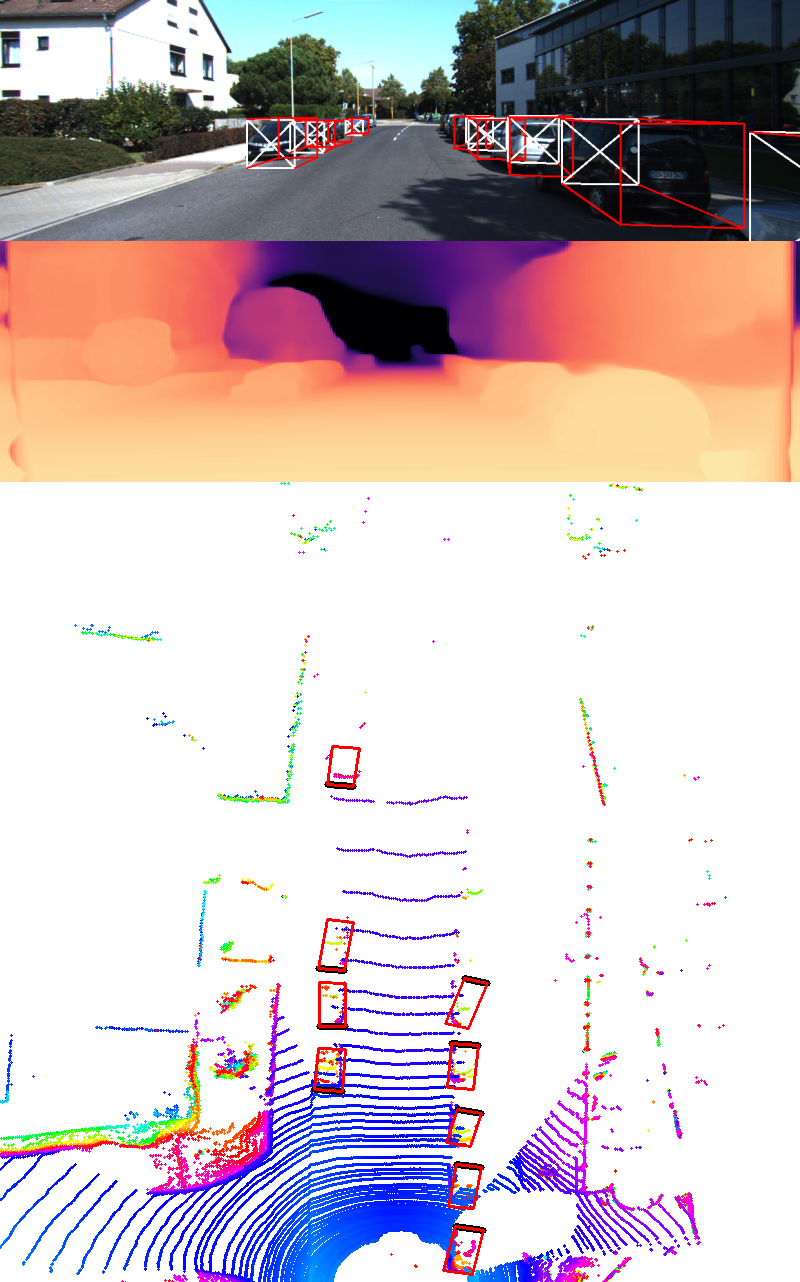}
  \caption{Qualitative results for joint LiDAR point-cloud based object detection and monocular depth prediction. Top row: Predicted 3D oriented bounding-boxes projected on left camera image plane, Middle row: Predicted dense depth map, Bottom row: Bird-eye-view of the predicted 3D oriented bounding-boxes overlayed on orthographic projection of LiDAR point-clouds (point-cloud is color-coded based on the height of each point).}
  \label{fig:qualitative_results}
\end{figure*}

We train our model with batch size of $8$, learning rate of $1.0 {e}^{-4}$, and \textit{Adam} optimizer for training loss minimization, on a PC with i9 10850K processor and a single NVIDIA RTX 3090 GPU which takes about $0.92$ seconds per step. With the KITTI object \textit{training} dataset split into $90\%$ \textit{train} and $10\%$ \textit{validation} samples, one epoch during training takes about $13$ minutes. Our complete network contains about $363.57M$ trainable parameters and we train them for $100$ epochs with \textit{early stop}, which takes about $22$ hours to train. During inference, VR3Dense runs at about $141.1fps$ on the same hardware. Figure \ref{fig:qualitative_results} shows few qualitative samples of VR3Dense prediction when tested on KITTI raw dataset \cite{Geiger2013IJRR}.

\subsection{Ablation Study}

% \begin{figure}[!ht]
% \centering
%   \includegraphics[width=0.95\linewidth]{fig/ablation_depth.png}
%   \caption{Ablation study to understand the effect of (a) adding supervision from the LiDAR points, (b) replacing the commonly used \textit{edge-aware smooth loss} with our \textit{edge-preserving smooth loss}, $\mathcal{L}_{eps}$, and (c) concatenating the feature vector learnt from voxel representation with the latent vector of monocular depth estimation network. L2: L2 loss between prediction and projected LiDAR points, S: Edge-Aware Smooth Loss, EPS: Edge-Preserving Smooth Loss, LC: Latent Concatenation (concatenation of object detection flattened features to the latent space).}
%   \label{fig:ablation_depth}
% \end{figure}

We study the effect of (a) adding supervision from LiDAR points, (b) replacing the commonly used \textit{edge-aware smooth loss} with our \textit{edge-preserving smooth loss}, $\mathcal{L}_{eps}$, and (c) concatenating the feature vector learnt from voxel representation with the latent vector of monocular depth estimation network; and we find that each of those contribute towards a better performance. We summarize these evaluations in table \ref{tab:ablation_study_depth}. For this evaluation, we only consider a subset of KITTI \textit{object} training dataset, where we randomly select a total of $1100$ frames and then divide them into $900$ \textit{train}, $100$ \textit{validation}, and $100$ \textit{test} samples. We then train our network with each of these configuration listed in table \ref{tab:ablation_study_depth} for $100$ epochs and notice a clear upward trend. 
%Qualitatively, test results after training the model on a small subset of dataset is shown in figure \ref{fig:ablation_depth}. 
Baseline model in the evaluations refer to our model with only \textit{unsupervised} loss, and as is evident from table \ref{tab:ablation_study_depth}, our model with an added \textit{supervision} from LiDAR points along with the \textit{edge-preserving smooth loss}, and voxel feature concatenation with  RGB-to-depth \textit{latent vector}, provides the best results. 
%In figure \ref{fig:ablation_depth}, this model seems to predict depth map with higher level of details and appears to be sharper than the baseline models. 

\begin{table*}[!htb]
    \vspace{-0.2cm}
    \begin{center}
      \caption{Ablation Study for Monocular Depth Estimation on a subset of KITTI object dataset. A subset of $1100$ data samples were obtained randomly from the \textit{training} dataset and divided into $900$ \textit{training}, $100$ \textit{validation}, and $100$ \textit{testing} samples. valuations are done on the pixels for which we have projected points from LiDAR available. L2: L2 loss between prediction and projected LiDAR points, S: Edge-Aware Smooth Loss, EPS: Edge-Preserving Smooth Loss, LC: Latent Concatenation (concatenation of object detection flattened features to the latent space).}
      \label{tab:ablation_study_depth}
      \begin{tabular}{l||l|l|l|l|l|l|l}
      \toprule
        \textbf{Method} & Abs Rel & Sq Rel & RMSE & RMSE log & $\delta < 1.25$ & $\delta < 1.25^2$ & $\delta < 1.25^3$ \\
        \midrule
         & \multicolumn{4}{c|}{\textit{Lower is better}} & \multicolumn{3}{c}{\textit{Higher is better}}\\ 
        \midrule
        \midrule
        Our Baseline                  & 0.414584 & 13.83466 & 12.28848 & 0.434995 & 0.680209 & 0.831007 & 0.904796 \\
        Our Baseline + L2             & 0.307075 & 6.505432 & 9.005099 & 0.368027 & 0.712899 & 0.861294 & 0.927120 \\
        Our Baseline + L2 + S         & 0.335025 & 6.186975 & 8.666602 & 0.383830 & 0.618205 & 0.855169 & 0.926737 \\
        Our Baseline + L2 + EPS       & 0.258648 & 3.151716 & 6.814891 & 0.327973 & 0.716325 & 0.865479 & 0.931353 \\
        Our Baseline + L2 + EPS + LC  & \textbf{0.248992} & \textbf{2.996437} & \textbf{6.667879} & \textbf{0.317969} & \textbf{0.724858} & \textbf{0.872039} & \textbf{0.935102} \\
        \bottomrule
      \end{tabular}
  \end{center}
\end{table*}

\section{Conclusion}

VR3Dense allows joint training of LiDAR point-cloud based 3D object detection and monocular depth estimation. We use 3D convolution layers for extracting spatial features from voxel representation of point-cloud and 2D convolution layers for extracting image features. Additionally, we concatenate the flattened spatial features with the latent vector of our RGB-to-depth encoder-decoder architecture and we show that it achieves a better performance compared to our baseline. While our object detection network is trained in a supervised way, we take a semi-supervised approach for depth estimation and further show that adding supervision from the 2D projection of LiDAR points helps the network converge faster. Lastly, we introduce a loss function, \textit{edge-preserving smooth loss}, which helps the network produce a sharper depth map with higher level of details, even when a \textit{pyramidal} structure is not present in the depth decoder for coarse-to-fine reconstruction of dense depth maps.

% \section*{Acknowledgments}

% This work was done as part of a coursework at Stanford University and we thank Prof. Silvio Savarese and Jeannette Bohg. Work done while working full-time as a Research Scientist at Ford Greenfield Labs, Palo Alto, CA.

\bibliographystyle{named}
\bibliography{vr3dense}

\end{document}